\pdfoutput=1

\documentclass[11pt]{article}

\usepackage{EMNLP2022}

\usepackage{times}
\usepackage{latexsym}

\usepackage[T1]{fontenc}

\usepackage[utf8]{inputenc}

\usepackage{microtype}

\usepackage{inconsolata}

\title{Gui at MixMT 2022 : English-Hinglish : An MT approach for translation of code mixed data}

\author {{Akshat Gahoi} \hspace{1cm} {\bf Jayant Duneja} \hspace{1cm}{\bf Anshul Padhi} \hspace{1cm} {\bf Shivam Mangale} \\ \hspace{1cm}{\bf Saransh Rajput}\hspace{1cm}{\bf Tanvi Kamble}\hspace{1cm}{\bf Dipti Misra Sharma} \hspace{1cm} {\bf Vasudeva Varma}  \\ \\
{ International Institute of Information Technology, Hyderabad} \\ \\
\{\href{mailto:akshat.gahoi@research.iiit.ac.in}{akshat.gahoi},\href{mailto:anshul.padhi@research.iiit.ac.in}{anshul.padhi},\href{mailto:saransh.rajput@research.iiit.ac.in}{saransh.rajput},\href{mailto:tanvi.kamble@research.iiit.ac.in}{tanvi.kamble}\}@research.iiit.ac.in  \\
\{\href{mailto:dunejajayant@gmail.com}{dunejajayant},\href{mailto:shivammangale@gmail.com}{shivammangale}\}@gmail.com }

\begin{document}
\maketitle
\begin{abstract}
Code-mixed machine translation has become an important task in multilingual communities and extending the task of machine translation to code mixed data has become a common task for these languages. In the shared tasks of WMT 2022, we try to tackle the same for both English + Hindi to Hinglish and Hinglish to English. The first task dealt with both Roman and Devanagari script as we had monolingual data in both English and Hindi whereas the second task only had data in Roman script. To our knowledge, we achieved one of the top ROUGE-L and WER scores for the first task of Monolingual to Code-Mixed machine translation. In this paper, we discuss the use of mBART with some special pre-processing and post-processing (transliteration from Devanagari to Roman) for the first task in detail and the experiments that we performed for the second task of translating code-mixed Hinglish to monolingual English. \\

\end{abstract}

\section{Introduction}

Code Mixing occurs when a multi-lingual individual uses two or more languages while communicating with others. It is the most natural form of conversation for multilinguals. It is often confused with code-switching but there is a slight difference between the two. Both these phenomena include communicating in multiple languages but code switching usually takes place within multiple sentences while code mixing usually refers to words of different languages used in the same sentence. In code mixing, phrases, words and morphemes of one language may be embedded within an utterance of another language. Code mixing is extensively observed on social media sites like Facebook and twitter. With the rapid growth of social media and consequently, increase in the use of code-mixed data, it becomes important to develop systems to process such text. \\
Machine Translation, also known as automated translation, is the process where a software translates text from one language to another without any human involvement. There are multiple forms of machine translation, however, over the past few years, neural machine translation has become extremely popular. 
The WMT shared task had two subtasks. The first subtask consisted of the translation of Hindi-English parallel sentence pairs to Hindi-English code mixed sentences through machine translation. The second subtask consisted of the translation of Hindi-English code mixed sentences to English. \\
 
 \section{Background}

 While there is a growing interest in code-mixed text analysis as a research problem, there is one bottleneck that has hindered the growth of such works, and that is the lack of data. Due to this, there aren’t many robust models for code-mixed text. To build standardized datasets of code-mixed text, we need to come up with ways of text generation of these code-mixed texts. These texts would be very helpful in training language models for various code-mixed pairs as language models only need unsupervised data. \\
 Code Mixed text generation is a relatively new problem, and so is its initial stage. One of the recent works in this field \cite{rizvi-etal-2021-gcm} tried to use linguistic theories to synthetically build code-mixed text using parallel monolingual corpora of two languages. The Equivalence Constraint Theory \cite{article} says that code-mixing can only occur at parts of the text where the surface structures of two languages map onto each other. So in these parts, the grammatical rules of both languages are followed. The Matrix Language Theory \cite{mcclure_1995} tries to solve this problem by separating the two languages into a base language and a second language. The grammatical rules of the base language are followed, and parts of the base language are replaced by the corresponding parts of the second language whenever it is grammatically feasible to do so. \\
 Deep Learning and Neural Networks have also been used to build systems for code mixed generation. In these systems, the problem of text generation has been posed as one of machine translation, where monolingual text is translated to code-mixed text. Some of the early work involved using the then state of the art encoder-decoder models like pointer generator  networks\cite{winata-etal-2019-code} and GANs\cite{chang19_interspeech} to translate two sets of monolingual corpora into code mixed text.
With the rise of multilingual models like mT5 \cite{https://doi.org/10.48550/arxiv.2010.11934}, mBART, indicBART \cite{https://doi.org/10.48550/arxiv.2109.02903}, etc. the task of translation has become much easier as these models understand both languages and this has been shown to outperform previous models in many workshops. \\
mBART\cite{https://doi.org/10.48550/arxiv.2001.08210} is a denoising autoencoder which has been trained on a very large dataset which contains text from 25 languages. It has the same transformer based architecture and training objective as BART, a denoising autoencoder which was shown to be one of the best performing sequence to sequence models at the time. It has been trained to reconstruct original text which has been corrupted as a way to add noise. It can perform various downstream sequence to sequence tasks like machine translation, text summarization, etc. mBART consists of 12 encoder layers and 12 decoder layers. There are 16 heads and a model dimension of 1024.  
\\
Another solution to circumvent the data problem is to create translation systems that can translate code-mixed text to monolingual text. This allows us to use robust NLP systems for various downstream tasks.\\
While we have the above said top performing models at the moment, they are very heavy computational wise due to their large parameter sizes. With resource constraints, it is tough to replicate their performance. Helsinki's OPUS-MT \cite{TiedemannThottingal:EAMT2020} model was of comparabaly smaller size and focused on the initiative of supporting low-resource languages. It does accordingly have lower performance. We have attempted at utilizing this model in our case with further training on provided data to understand whether under the resource constraints, we can observe competitive performance.
\\
The model architecture is based on a standard transformer setup with 6 self-attentive layers in the encoder and decoder network. It has 8 attention heads in each layer. This is hence comparatively low compute seeking as compared to the mainstream models.

\section{System Overview}

\subsection{Task 1}

In this section we propose our system for Task 1 which is English and Hindi to code-mixed text translation

\subsubsection{Dataset and Data Preparation}
The dataset that we used for Task 1 was the HinGE dataset \cite{inproceedings}. It is divide into two parts, the synthetic dataset or the machine generated dataset and the human generated dataset. (Table 1) There were 3659 and 2176 sentences respectively.
\begin{table}
\centering
\begin{tabular}{lc}
\hline
\textbf{Data} & \textbf{Length}\\
\hline
\verb|Synthetic (Train)| & {3263} \\\hline
\verb|Synthetic (Validate)| & {396} \\\hline
\verb|Human Generated (Train)| & {1800} \\\hline
\verb|Human Generated (Validate)| & {376} \\\hline
\end{tabular}
\caption{Distribution of Sentences in the data}
\label{tab:accents}
\end{table}

\subsubsection{Model}
In this task we finetune the mBART model on the data given to us. Since mBART is a very large model we needed to decrease its size. We do this by reducing the vocabulary of the model as the vocabulary adds to the model size by a lot and we don’t need the vocabulary from the rest of the 25 languages. To reduce the vocabulary we create our own vocabulary using the tokens present in the task dataset, IIT-B English-Hindi parallel corpus \cite{kunchukuttan-etal-2018-iit} and the Dakshina Dataset \cite{https://doi.org/10.48550/arxiv.2007.01176} as we feel the two datasets were large enough to create a vocabulary extensive enough to solve the given task. We process the input data from the given task data as explained above to create our input. Using the corresponding code-mixed sentences as the gold output we finetune the mBART model.
\begin{table*}
\centering
\begin{tabular}{lll}
\hline
\textbf{Post Processing} & \textbf{ROUGE-L} & \textbf{WER}\\
\hline
Normal Output & 0.39091 & 0.81884 \\
With Automated Transliteration & 0.48376 & 0.72561 \\
\textbf{With Automated Transliteration + Dictionary Based Transliteration} & \textbf{0.61667} & \textbf{0.63342} \\
\hline
\end{tabular}
\caption{\label{citation-guide}
ROUGE-L and WER scores after different post processing tasks
}
\end{table*}

\subsubsection{Post Processing}
The output of our model was in a mixed script (Roman + Devanagari). So the post processing becomes one of the important step in this task as we wanted our output Hinglish sentences to be only in Roman script. We used transliteration function from indicate library as our first step to see how good the results will be. There were many instances where the transliteration done by indicate was not accurate. So the next step that we did was to create a dictionary of most common words and numbers with their corresponding transliterated Roman text. This dictionary over the automatic transliteration by indicate was used to get the best output of our model in the Roman script.

\subsection{Task 2}
In this section we propose our system for Task 2 which is Hinglish (code-mixed) to English text translation.

\subsubsection{Observations}
The data for task 2 are tweets based data. Due to the tweets nature, we observed that:
\begin{itemize}
    \item The URLs included tended to be at the end of the sentences.
    \item The mentions (of the form '@<some\_user\_tag>' for instance @LokSabha) at the beginning and the end of tweets are generally such that the sentences can be translated without them with no-low loss of information.
    \item Hashtags which are added at the end of the tweets are generally for increasing outreach and exposure.
\end{itemize}

Based on the above observations, we found that the information provided by these tokens to the translation was not significant as compared to the loss of information due to incorrect translation of these units. Hence, we applied heuristics to appropriately preprocess the input data to exclusively and exhaustively split the tweets into sentences (which will be translated), URLs, mentions and hashtags, which are then concatenated after the translation in postprocessing.

\subsubsection{Dataset and Data Preparation}
The dataset that we used for Task 2 was the PHINC dataset \cite{srivastava-singh-2020-phinc}. It contains 13,738 parallel sentences in Hinglish (code-mixed) and English of which we used a train-val-test split of 80-10-10. We transliterated the Hinglish sentences from the Roman script to the Devanagiri script using the  Google Transliterate API, to utilise pre-trained Hindi to English translation models. This transliterator was used among others due to it having one of the best performance, it's similarity in the vocabulary space with the input dataset as compared to the other transliterators available and also that PHINC was jointly created using Google Translate.\label{sec:data_task_2}

\subsubsection{Model}
Due to compute constraints, we decided to utilize pretrained models, that would be efficient for our dataset. To access better models, we went ahead with models trained with a task or a subtask of Hindi to English machine translation. We appropriately processed the data for the same. We hence decided to finetune Salesken.AI's pretrained model provided on Huggingface Transformers. They have finetuned Helsinki's OPUS-MT model on AI4Bharat's Samanantar dataset \cite{ramesh2021samanantar}, a large indic dataset.

\section{Experimental Setup}
In task 1, we use the fairseq implementation of mBART as our base model which has been trained on 4 Nvidia GeForce RTX 2080 Ti GPUs. The model has been trained using label smoothed cross entropy as the loss criterion.
The model uses an Adam optimizer with polynomial decay learning rate scheduling, dropout = 0.3, learning rate = $3*10^{-5}$, $\epsilon$ = $10^{-6}$, $\beta_1$ = 0.9 and $\beta_2$ = 0.98.

The model was trained on 10000 steps with 2500 warm up steps and a batch size of 512 tokens.

We validate the model on each epoch on a validation set and at the end we select the model with the lowest loss.

In task2, for fine-tuning we use the Salesken.AI’s pretrained model provided on Huggingface Transformers. The model was trained on Nvidia GeForce RTX 2080 Ti GPUs.The model has been trained using label smoothed cross entropy as the loss criterion. The model uses an Adam optimizer with learning rate = $3*10^{-4}$, $\epsilon$ = $10^{-9}$, $\beta_1$ = 0.9 and $\beta_2$ = 0.98.

\section{Results and Evaluation}
The test dataset consisted of 500 sentences. These sentences also had both English sentence and its corresponding Hindi sentence. ROUGE-L score and WER score was considered for evaluation. ROUGE-L score considers longest common subsequence for its scoring. It counts the longest subsequence which is shared between both reference and the output. Its different from precision as it only counts the ratio between longest subsequence matched and the number of words matched. It does not take all the words in the reference. \\
The WER score represents the word error rate. Total errors between the reference and output is considered for this score. It adds up all the substitution, addition and deletion required to convert the output to the reference sentence and treat it as total error of the output. It can be treated same as calculating Levenshtein distance. \\
So our aim was to maximise ROUGE-L score and minimize WER score. Our score improved as we translitered the output from Devanagari to Roman using indicate library. The score increase significantly after we created a dictionary of words for transliterating most common Hindi words and numbers. We achieved a ROUGE-L score of 0.61667 and WER score of 0.63342 after both the post processing steps.

The test set provided for Task 2 contained 1500 lines, which were processed as mentioned in \ref{sec:data_task_2}
The results for the evaluation metrics we obtained for the test set provided for Task 2 is available in \ref{tab:results_task2}. Using the Google Transliterate API significantly improved the quality of the input data, and also the similarity of vocabulary with the dataset as mentioned earlier. \label{sec:data_task_2} The application of heuristics also bolstered the approach's performance.

Based on qualitative evaluation, it was observed that it struggled to get long sentence translations which can be attributed to the source of the dataset being of of tweets which have a noisy and inconsistent structure. This is alongside the lower parameter size and attention heads.

The model was trained till significant learning on a wide array of parameters, till resource permits, in an attempt to provide more opportunities to appropriately fine-tune the model, but even though there was a sign of the model learning, the performance was observed to be not competitive to the current top performers.

\begin{table}
\centering
\begin{tabular}{lc}
\hline
\textbf{Metric} & \textbf{Score}\\
\hline
\verb|ROUGE-L| & {0.41493} \\\hline
\verb|WER| & {0.80804} \\\hline
\end{tabular}
\caption{Results for Task 2}
\label{tab:results_task2}
\end{table}

\section{Conclusion}
In this paper, we approached code-mixed machine translation problem from both the direction. We used mBART for our first task of translating English and corresponding Hindi sentences to Hinglish sentence. The results were significantly improved through transliterating the output from Devanagari script to Roman script. Two different methods were used for the same. Our model surpassed baseline in ROUGE-L and WER scores by a huge margin. \\
For the second task of translating Hinglish sentences to English sentence by fine-tuning Salesken.AI's pre trained model. We cleared the baseline but there is still work to be done in that field as we think that it can be further improved. For future work in this area, we would like to work further on the second task in hand of translating codemixed language to a monolingual language. We need to retrieve information about both languages from the code mixed sentence and try to give an output in a mono-lingual langauge without disturbing the word order.

\bibliography{anthology,custom}
\bibliographystyle{acl_natbib}

\end{document}